\RequirePackage{fix-cm}
\documentclass[smallextended]{svjour3}       
\pdfoutput=1
\smartqed  

\usepackage{mathptmx}       
\usepackage{helvet}         
\usepackage{courier}        
\usepackage{type1cm}        
\usepackage{makeidx}         
\usepackage{graphicx}        
\usepackage{multicol}        
\usepackage{booktabs}
\usepackage[bottom]{footmisc}
\usepackage{paralist}
\usepackage{tikz}

\makeindex             

\newcommand{\notaB}[1]{}

\newcommand{\hy}{\hat{Y}}
\newcommand{\posout}{1}
\newcommand{\nposout}{0}
\newcommand{\possa}{1}
\newcommand{\npossa}{0}

\begin{document}

\title{Discrimination in machine learning algorithms}
\author{Roberta Pappad\`a and Francesco Pauli}
\institute{Department of Economics, Business, Mathematics and Statistics ``B. de Finetti'' , University of Trieste, Trieste, Italy, \email{rpappada@units.it, francesco.pauli@deams.units.it} }
%
%
\date{Received: date / Accepted: date}

\maketitle

\begin{abstract}
Machine learning algorithms are routinely used for business decisions that may directly affect individuals, for example, because a credit scoring algorithm refuses them a loan.
It is then relevant from an ethical (and legal) point of view to ensure that these algorithms do not discriminate based on sensitive attributes (like sex or race), which may occur unwittingly and unknowingly by the operator and the management.
Statistical tools and methods are then required to detect and eliminate such potential biases.
\end{abstract}

\section{Introduction}
\label{sec:intro}
The kind of discrimination we refer to consists of treating a person or a group depending on some sensitive attribute (s.a., $S$) such as race (skin color), sex, religious orientation, etc. 

\begin{figure}
	\begin{center}
		\begin{tikzpicture}[node distance=0.8cm]
		\node (a) at (0:1) {{\it (a)}};
		\node (s) [below of = a] {$S$};
		\node (Y)  [below of = s] {$Y$};
		\path (s) edge [->]   (Y);

		\node (b) at (0:3) {{\it (b)}};
		\node (s) [below of = b] {$S$};
		\node (X1)  [below of = s] {$Z$};
		\node (Y)  [below of = X1] {$Y$};
		\path (s) edge [->]  (X1);
		\path (X1) edge [->]  (Y);

		\node (c) at (0:5) {{\it (c)}};
		\node (s) [below of = c] {$S$};
		\node (X1)  [below of = s] {$X_1$};
		\node (Y)  [below of = X1] {$Y$};
		\path (s) edge [->]  (X1);
		\path (X1) edge [->]  (Y);

		\node (d) at (0:7) {{\it (d)}};
		\node (s) [below of = d] {$S$};
		\node (X1)  [below left of = s] {$X_2$};
		\node (Y)  [below right of = X1] {$Y$};
		\path (s) edge [->]  (X1);
		\path (s) edge [->]  (Y);

		\node (e) at (0:9) {{\it (e)}};
		\node (s) [below of = e] {$S$};
		\node (X1)  [below left of = s] {$X_2$};
		\node (Z)  [below right of = s] {$Z$};
		\node (Y)  [below  of = Z] {$Y$};
		\path (s) edge [->]  (X1);
		\path (s) edge [->]  (Z);
		\path (Z) edge [->]  (Y);

		\node (f) at (0:11) {{\it (f)}};
		\node (s) [below of = f] {$S$};
		\node (X1)  [below left of = s] {$X_2$};
		\node (Z)  [below right of = s] {$Z$};
		\node (Y)  [below  left of = Z] {$Y$};
		\path (s) edge [->]  (X1);
		\path (s) edge [->]  (Z);
		\path (s) edge [->]  (Y);
		\path (Z) edge [->]  (Y);
	\end{tikzpicture}
	\end{center}
	\caption{\label{fig:dags} $Y$ is the outcome, $S$ is the sensitive attribute, $X_i$ denotes observed variables, $Z$ denotes an unobserved variable. Example: $S$: race, $Y$: restitution of a loan, $X_1$: socioeconomic status, $X_2$: zip code residence, $Z$ availability of family financial support.}
\end{figure}
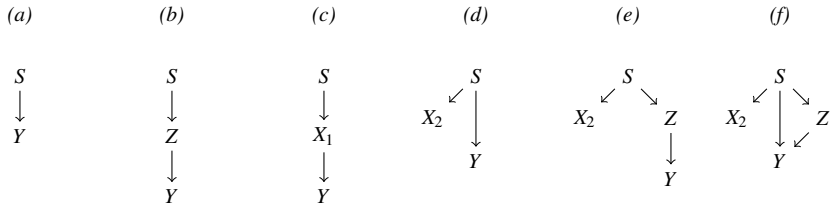

A human may discriminate either because of irrational prejudice induced by ignorance and stereotypes or based on statistical generalization: lacking specific information on an individual, he is assigned the characteristics prevalent in the sensitive attribute category he belongs to. For example, in the United States, lacking information on education, a black person may be assumed to have relatively low level since this is the case in general for black people in the country) \cite{romei2014multidisciplinary}.
When a statistical or machine learning algorithm is used in the decision process, its behavior concerning  discrimination depends on the information it is given. 
In particular, if the sensitive attribute is available to the algorithm (i.e., it is included in the learning data and can be used for predictions), it may discriminate either because the data it is taught contain irrational prejudice (Fig.\,\ref{fig:dags}{\it (a)}) or because the sensitive attribute is associated to an unobserved attribute that is relevant for the prediction of $Y$, the outcome of interest (Fig.\,\ref{fig:dags}{\it (b)}).
An example of the former may occur if an algorithm is sought for screening job applicants, and the learning data consist of selections made in the past by humans who, in some instances, based their decisions on irrational prejudice, consciously or not. As an example of the latter, consider a credit scoring algorithm where the applicant's educational attainment  is unknown but the race is, then because education is relevant to assess the applicant's reliability and is associated to race, the latter may be used in the decision. 
We note in passing that, although this is a rare circumstance, it may also be the case that a sensitive attribute is directly related to the outcome not because of irrational human prejudice, as it happens in life insurance for the gender attribute.

If the algorithm is constrained not to use the s.a. $S$ in the final rule, i.e., is excluded from the data from which the algorithm learns, discrimination could still happen either because a variable that is related to the outcome is also related to $S$ (Fig.\,\ref{fig:dags}{\it (c)}) or because a variable included in the data that is unrelated to the outcome is related to $S$, which is associated with the outcome either directly as in Fig.\,\ref{fig:dags}{\it (d)}, mediated  as in   Fig.\,\ref{fig:dags}{\it (e)},  or both ways as in  Fig.\,\ref{fig:dags}{\it (f)}).
In the first case, the collective possessing the s.a. may experience the desired outcome less frequently than the rest whenever it differs from the population in the distribution of some relevant features  in the data. 
In the second case  the collective possessing the s.a. may experience the desired outcome less frequently than the rest even if it were equal to the population as far as the other relevant features included in the data are concerned. In this second situation, the discrimination may be due to a difference in the distribution of some relevant feature not included in the data but related to the s.a. and some variable in the sample.

Whether or not the situations described above are instances of undesired discrimination depends on how the concept is defined from a legal or ethical standpoint. 
In particular, it depends on whether we require that the groups identified by $S$ have the same treatment (rate of positive outcome) unconditionally or that they have the same treatment conditionally on some relevant characteristics different from $S$ and are deemed lawful to use to discriminate. 

Which definition is more appropriate is a matter of ethic or law. A requirement of unconditional equal treatment means that the groups must be given equal treatment even if they are not equal (demographic parity, disparate impact), which may seem unjust from an individual point of view \cite{hardt2016equality}. 
On the other hand, admitting the groups to be treated differently because one of them possesses desirable characteristics possibly amounts to perpetuating past unfair discrimination, which may have led the groups to be different.
The first requirement may be an instance of affirmative action since it goes in the direction of eliminating the differences between groups, differences which are at least partly admissible within the second requirement.
If conditional equal treatment is sought, one must decide which characteristics other than S but possibly correlated with it are ethical/lawful to use, which may be partly dictated by law, partly uncertain (for example in a civil law system such as in the US whether a characteristic is lawful to discriminate on may be for a jury to decide).

In order to give precise definitions, from now on assume both $S$ and $Y$ dichotomous, $Y=1$ be the desired outcome and $S=1$ the belonging to a protected category. 
If unconditional equal treatment (demographic parity) is desired, data are apt if it is not possible to predict $S$ from $Y$ \cite{feldman2015certifying}, that is, data are compatible with
\[ P(Y=\posout|S=\possa) = P(Y=\posout|S=\npossa). \]

If equal treatment should be conditional on $X_1$ then the requirement becomes
\[ P(Y=\posout|S=\possa, X_1=x_1) = P(Y=\posout|S=\npossa, X_1=x_1). \]

An extreme version of conditional equal treatment is advocated in \cite{hardt2016equality}, a rule is non discriminatory if the prediction errors are the same regardless of $S$
\[ P(\hy=\posout|S=\nposout, Y=y ) = P(\hy=\posout|S=1,Y=y),\;\;\forall y;\]
or, which is the same, prediction and sensitive attribute are independent conditional on the outcome. 

\section{Measuring and avoiding discrimination by causal inference}
\label{sec:lit}

The different strategies which can be used to avoid discrimination from an algorithm can be distinguished depending on the level at which action is taken: 1) learning data can be modified to ensure they do not imply discrimination; 2) the learning algorithm can be integrated with a non discrimination objective; 3) the final algorithm can be tampered with after it has been fitted; 4) the final predictions can be changed. 
The precise action to be taken depends on the definition of discrimination adopted; as outlined in the previous section, we may seek unconditional equality or conditional equality. 

We focus on data preprocessing techniques.
In a nutshell, this entails first establishing whether and to what extent the available data are discriminatory.
If discrimination is detected data are modified in order to make them discrimination free before and then used with a standard algorithm \cite{kamiran2012data}. 
Various ways of modifying data have been proposed including: 1) removing attributes (correlated with $S$); 2) changing the labels of some units; 3) weighing units; 4) resampling units
We note that the precise implementation for both steps depends on the definition of discrimination which is adopted, presumably on legal/ethical grounds (see section \ref{sec:intro}). 


In order to measure discrimination within a dataset it has been proposed to use causal inference techniques. 
Causal inference techniques aim at estimating the causal effect of a treatment--in this context, the belonging to a protected category--on an outcome. 
Generally speaking, causal inference methods aim at assessing the effect of a treatment based on observational data by matching treated units to untreated units which are similar with respect to their observed characteristics: a balanced dataset, suitable to draw causal inference, is built by restricting the original dataset to treated and untreated units which have been matched.
Comparing a protected units outcome with the outcome of unprotected units which are similar with respect to the other observables is also a reasonable way to detect whether the unit has been discriminated (conditionally).

Following this idea, Luong {\em et al.} \cite{luong2011k} propose to measure discrimination for unit (individual) $i$ of the protected category ($S_i=1$) as
\begin{equation}
\label{eq:delta}
\Delta_i = \frac{1}{k} \#\{j|j\neq i, x_j\in U^{k,1}_i, y_j=y_i\}
-
\frac{1}{k} \#\{j|j\neq i, x_j\in U^{k,0}_{i}, y_j=y_i\}
\end{equation}
where $U^{k,s}_{i}$ is the set of the $k$ nearest neighbours of $x_i$ within those units for which $S=s$, according to a Gower type distance (possibly, other measure of the difference between the two frequencies are used such as the ratio or the odds). Note that a positive $\Delta$ indicates discrimination against the protected category if $y_i$ is the undesired outcome or discrimination in favour of the protected category if $y_i$ is the desired outcome. The authors suggest fixing a threshold $\tau\in [0,1]$ to declare the individual $i$ discriminated if $\Delta_i\geq \tau$, where $\tau$. 
To prevent discrimination the dataset is changed by altering the value of $y_i$ for those units fo which $\Delta_i>\tau$. $\tau$ is then a tuning parameter which regulates the trade off between residual discrimination and accuracy.

In what follows we focus on discrimination against the protected group, and so we compute a different version of the measure $\Delta$:
\begin{equation}
\label{eq:deltaprimo}
\delta_i = \frac{1}{k} \#\{j|j\neq i, x_j\in U^{k,1}_i, y_j=0\}
-
\frac{1}{k} \#\{j|j\neq i, x_j\in U^{k,0}_{i}, y_j=0\}
\end{equation}

\section{CEM based discrimination measure}

Coarsened Exact Matching (CEM, \cite{iacus2011multivariate,iacus2009cem}) is based on coarsening continuous variables and match a treated unit to those untreated units which are equal with respect to the coarsened continuous variables and the categorical ones.
In order to obtain a reliable estimate of the causal effect, CEM algorithm discards those units which can not be matched.
Here, the objective is different in that we are not interested in a global estimate of the effect of the S, but rather whether unit $i$ has experienced a different outcome because of it possesses the S: a comparison of unit $i$ outcome with the outcome of units matched by CEM is a suitable measure of discrimination.
In particular, let $\bar{y}_i^{(S=0)}$ be the relative frequency of positive outcomes among units matched with unit $i$ (which possesses the s.a.) and not possessing the s.a., then $D_i=y_i-\bar{y}_i^{(S=0)}$ is a measure of discrimination which takes negative values when a unit is discriminated against and positive values when the unit is favoured (positive discrimination), so that the value of $D_i$ is bounded between $-1$ and $1$. 
However, in a standard implementation of CEM algorithm not all units are matched, which makes it unsuitable for our purpose.
A possible strategy to exploit CEM technique is to apply it sequentially as follows. 
Suppose that all units are matched using $k$ variables, it is possible that once an additional variable is considered in matching some units remain unmatched, we then measure the discrimination for those units based on the matching with $k$ variables. 
Starting with an initial matching on $0$ variables, that is, $D_i^{(0)}=y_i-\bar{y}^{(S=0)}$ where $\bar{y}^{(S=0)}$ is the relative frequency of the positive outcome for all units in the dataset not possessing the s.a., the sequential CEM allows to obtain a discrimination measure for all units. Specifically, assuming that a non empty subset $C_i$ of similar units can be found based on $k$ variables, the discrimination against unit $i$ is measured as follows
$$
D_{i}^{(k)}= y_i - \frac{ \#\{j|y_j=1,s_j=0,j\in C_i\}}{\#\{j|s_j=0,j\in C_i\}},
$$	
for $j\neq i$, which  takes negative values when the unit is discriminated against.
Note that, differently from the procedure proposed in \cite{luong2011k}, the units are similar when they share the same values on the available variables, possibly after coarsening the continuous ones. Moreover, the sequential CEM procedure returns a measure $D_i$ for each unit (at worst, it is based on the rate of positive outcome for the whole non sensitive sample).  
Clearly, the discrimination measures $D_i$ depend on the order of addition of the variables, hence the procedure is repeated for different (random) orders of addition and the final result is the average (See Fig.\,\ref{fig:algorithm}). Since computing the measure for all possible orders is typically non feasible, the number of repetitions is decided based on convergence of the $D_i$ measures.


\begin{figure}
	\fbox{
	\parbox[t]{0.97\textwidth}{
	Let 
	\begin{compactitem}
		\item $x_1,\ldots,x_K$ be the available variables; 
		\item let $\mathcal M^{(j_1,\ldots,j_h)}$ be the set of units matched by CEM performed using variables $x_{j_1},\ldots,x_{j_h}$ and let $C^{(j_1,\ldots,j_h)}_i$ be the set of units belonging to the same CEM cell of unit $i$.
	\end{compactitem}
	Repeat $M$ times
	\begin{enumerate}
		\item[$-1$:] select a random permutation $i_1,\ldots,i_K$ of $1,\ldots,K$;
		\item[0:] for all units $i$ let $D_i^{(0)}=y_i-\bar{y}$ where 
		$\bar{y}= \hat{P}\{Y=1|S=0\} = \#\{i|y_i=1,s_i=0\}/\#\{i|s_i=0\}$;
		\item[$k$:] 
		\begin{itemize}
			\item for all $i\notin \mathcal M^{(i_1,\ldots,i_k)}$ let $D_i^{(k)}=D_i^{(k-1)}$;
			\item for all $i\in \mathcal M^{(i_1,\ldots,i_k)}$ let $D_i^{(k)}=y_i-\bar{y}_i$ where $\bar{y}_i= \hat{P}\{Y=1|S=0,x_{i_1},\ldots,x_{i_k}\} = \#\{i|y_i=1,s_i=0,i\in C^{(i_1,\ldots,i_k)}_i\}/\#\{i|s_i=0,i\in C^{(i_1,\ldots,i_k)}_i\}$;
		\end{itemize}
		\item[$K+1$:] $D_i^{(i_1,\ldots,i_k)}=D^{(K)}_i$.
	\end{enumerate}
	Set the final discrimination scores as
	$ D_i=\frac{1}{M} \sum_{i_1,\ldots,i_K} D_i^{(i_1,\ldots,i_k)} $.}}
	\caption{\label{fig:algorithm} Pseudo-code for repeated sequential implementation of CEM.}
\end{figure}

\section{Testing the procedure}

Discrimination measures can be tested by assessing whether individuals have a positive outcome more often if they do not belong to the protected group conditional on $\bf X$, the set of `neutral' characteristics (different than S), which are deemed lawful to discriminate upon.
In order to explore the performance of the proposed measure of discrimination we consider three different scenarios: (i) discrimination free data; (ii) discriminating data; (iii) presence of a variable that is related to $S$ but unrelated to the outcome $Y$.

As mentioned before, the discrimination measure $D_i$ takes values in $[-1, 0)$, then this means that discrimination against the $i$-th individual has been detected, as the value $y_i$ is less than the relative frequency of positive outcomes among units matched with $i$ (having the s.a.) and not possessing the s.a., $\bar{y}_i^{(S=0)}$. By contrast, positive values of $D_i$ may indicate that discrimination in favour of individual $i$ has occurred.  

For comparison purposes, we consider the measure $\delta$ in Eq.(\ref{eq:deltaprimo}), using $32$ as $k$ value, where $\delta_i>0$ indicate that the difference in the rate of unfavourable outcomes in sensitive units and non sensitive ones among the $k$ nearest neighbours of unit $i$ is positive, thus providing a measure of discrimination against a protected category under the $k$-nn approach (hereafter, KNN). 

In what follows, we first present the data used to test the proposed discrimination measure. Then, different strategies to check the stability of the procedure and its performance in detecting and measuring discrimination against a protected group are discussed. Finally, the results from simulation studies and the comparison with the $k$-nn approach are illustrated. 
 
\subsection{The data}

\paragraph{The adult dataset.}

The {\it adult} dataset \cite{luong2011k} is related to people income based on census data.  The data contains 45222 observations and attributes  age, years of education, working hours, type of occupation (sector, type of employer), family status, sex, capital gain and loss, native country (redefined as areas of the world), ethnicity and income. Here, rather than a decision, the outcome considered in our analysis is having an income lower or equal than \$50K, and higher than \$50K. As sensitive attribute ($S$) we consider being non white, with the aim of investigating the possible presence of social discrimination against people in the protected group. Hence $Y=1$ is the desired outcome of having an income greater than $\$50\,000$ (corresponding to the $28.76\%$ of units).  For the analysis, the dataset is split in a learning (30162 units) and test (15060 units) subsamples.

\paragraph{The COMPAS dataset.}
The {\it COMPAS} dataset is part of the public database of  Parolees Under-Community-Supervision for the State of New York. The data refer to years 2018 and 2019 and are based on COMPAS supervision levels. COMPAS (Correctional Offender Management Profiling for Alternative Sanctions) is a popular algorithm used by Judges and Parole Officers for scoring criminal defendant's likelihood of reoffending (recidivism). The COMPAS model requires that, based on the risk level, each inmate is assigned a score, and consequently a level of supervision among four possible levels: level 1 is the highest level of supervision and at most risk, requiring 4 face to face contacts per month. Levels 2 and 3 require 3 face to face contacts per month and 2 face to face contacts every three months. Finally, COMPAS level 4 cases require 2 face to face contacts every four months, being the lowest level of supervision. In order to test the CEM-based discrimination measure we analyse 40704 observations and include the year represented by the snapshot population, the Region of supervision, the County of residence, gender, age, and type of crime. The protected group consists of non-white people (61.8\% units), and the desired outcome ($Y=1$) is the lowest COMPAS supervision level, which is possessed by about 28.75\% inmates.  For the analysis, the dataset is split in a learning (32488 units) and test (8216 units) subsamples. 

\paragraph{The Custody dataset.}

The {\it Custody} dataset is hosted by the Department of Correction Agency of the City of New York and updated daily. The data cover the period 2016--2019 (7458 observations) and concern daily inmates in custody with information on the date and time of the incident, mental health designation, ethnicity, gender, age, legal status (whether an inmate ID a detainee), security risk group membership, top charge for the inmate, and infraction flag. This dataset excludes Sealed Cases. The custody level is a categorical variable with levels MIN, MED, MAX. We set the s.a. being non white (possessed by $88.29\%$ of units) and positive outcome being the minimum custody level (MIN), which is present in 2355 cases ($31.58\%$). The data are available at https://data.cityofnewyork.us/Public-Safety/Daily-Inmates-In-Custody/7479-ugqb/data.

\subsection{Removing and adding discrimination}
\label{sec:discadd}
To simulate a sample where no discrimination is present, the following alternative strategies are adopted:
\begin{itemize}
	\item[(a)]  the outcome is simulated for all units according to the estimated probabilities from the pruned classification tree;
	\item[(b)] the outcome is simulated for all units according to the predicted values without relying on pruning the tree;
	\item[(c)] the observations in $S$ are modified by means of a random permutation on the units for which the classification tree estimate of the  probability of a positive outcome is within certain quantiles; 
    \item[(d)] a random permutation is performed on the observations of the sensitive attribute using all the units in the dataset.
\end{itemize}
Note that in (a) and (b) discrimination is eliminated by modifying the outcome $Y$ in the dataset, whereas strategies (c) and (d) build a discrimination free dataset by modifying the sensitive attribute $S$ for either all or a subset of units. 

Further, we modify the dataset by adding  discrimination against the protected class to assess the sensitivity of the discrimination scores. In particular, we introduce discrimination according to the following approaches:
\begin{enumerate}
\item the outcome is changed to negative for a certain percentage of units having the s.a.\ and a positive outcome; 
\item the outcome is modified to positive for a certain percentage of units which are not in the protected group and have a negative outcome. 
\end{enumerate}
The units for which we modify the outcome are chosen among those for which the probability of positive outcome is within a given range, depending on fraction of observations changed. Different scenarios arise from the combination of the two strategies. In particular, let $v_1$ be the percentage of units $y_i$ with $S=1$ (i.e., having the s.a.) for which the outcome is changed from $Y=1$ to $Y=0$, and let $v_2$ be the percentage of units $y_i$ with $S=0$ (not possessing the s.a.) for which the outcome is changed from $Y=0$ to $Y=1$. By letting $v_1>0$, we increase the amount of discrimination against the protected category, whereas a positive $v_2$ allows to introduce discrimination in favour of the unprotected category.

\begin{figure}[h]
	\begin{center}
		\includegraphics[width=.98\textwidth]{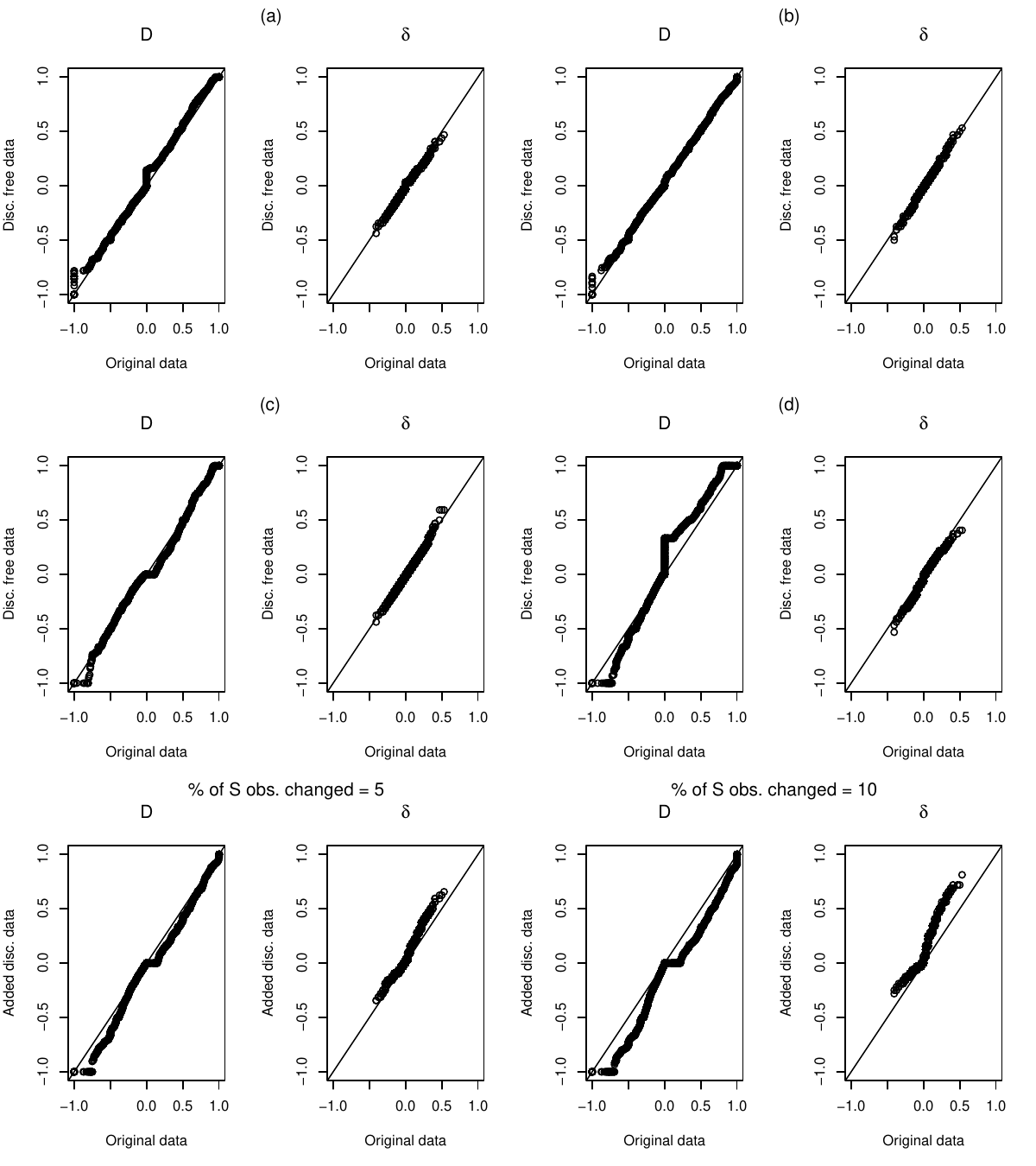} 
	\end{center}	
	\caption{\label{fig:qqplot_adult} Discrimination measures  performances for the \emph{adult} data: qq-plots comparing the distributions of the discrimination scores $D$ and $\delta$ for (i) the discrimination free data against the original data according to strategies (a)--(d) described in Sect. \ref{sec:discadd} (first two rows from the top) and (ii) the data where discrimination has been introduced against the original data, by setting $p_1=p_2=5$  and  $p_1=p_2=10$ (bottom row).}
\end{figure}

We compare the distributions of the discrimination scores returned by the algorithm of Fig.\ref{fig:algorithm} /with 100 iterations) using $qq$-plots to assess how well the different measures are able to detect the simulated scenarios. Figure \ref{fig:qqplot_adult} shows the $qq$-plots for the \emph{adult} dataset and the comparison of the original data against the discrimination free data (simulated) according to the methods (a)--(d) described above.

\begin{figure}[h]
	\begin{center}
		\includegraphics[width=.98\textwidth]{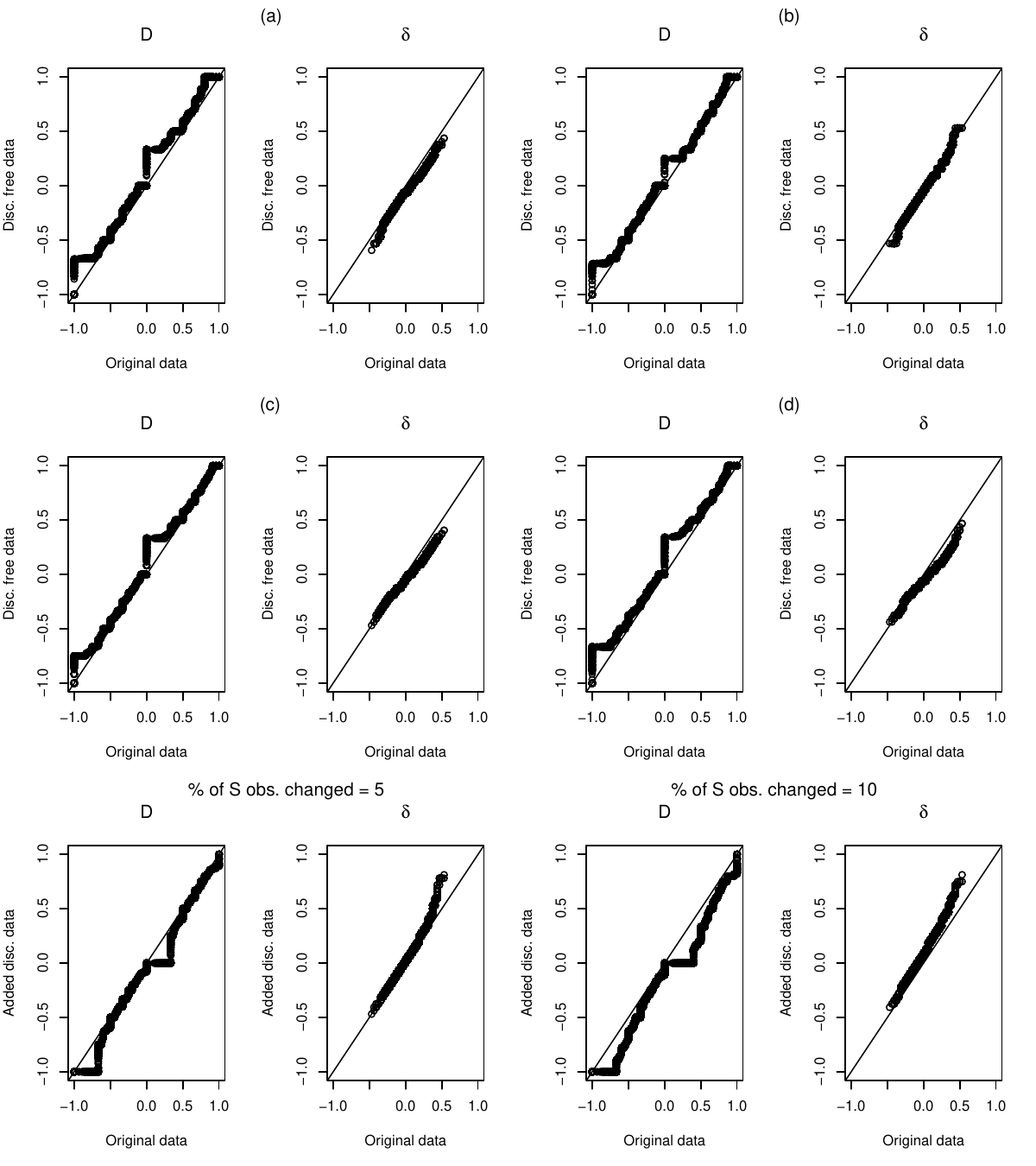} 
	\end{center}	
	\caption{\label{fig:qqplot_compas} Discrimination measures  performances for the \emph{COMPAS} data: qq-plots comparing the distributions of the discrimination scores $D$ and $\delta$ for (i) the discrimination free data against the original data according to strategies (a)--(d) described in Sect. \ref{sec:discadd} (first two rows from the top) and (ii) the data where discrimination has been introduced against the original data, by setting $p_1=p_2=5$  and  $p_1=p_2=10$ (bottom row).}
\end{figure}

With respect to $D$, the measure $\delta$ shows less variability regardless of the method used. Indeed, from scenarios (a)--(d) it can be seen that $D$ is the only one where a difference is spotted, with negative values indicating discrimination against the protected category. Moreover, the last row of Figure \ref{fig:qqplot_adult} displaying the comparison between the original and the discriminating data (simulated) confirms that the proposed measure $D$ is able to detect the introduced discrimination, yielding discrimination scores that are lower than that for the raw data when $v_1=v_2=5$ and to a larger extent when $v_1=v_2=10$. Note that the computation of the measure $\delta$ is always based on groups of fixed size $k$, while very small (CEM) strata may contribute to the values of $D$, hence $\delta$ may be overall less sensitive to detect discrimination.
Similar considerations can be drawn from \ref{fig:qqplot_compas} for the \emph{COMPAS} dataset.

\begin{figure}[h]
\begin{center}
\includegraphics[width=.98\textwidth]{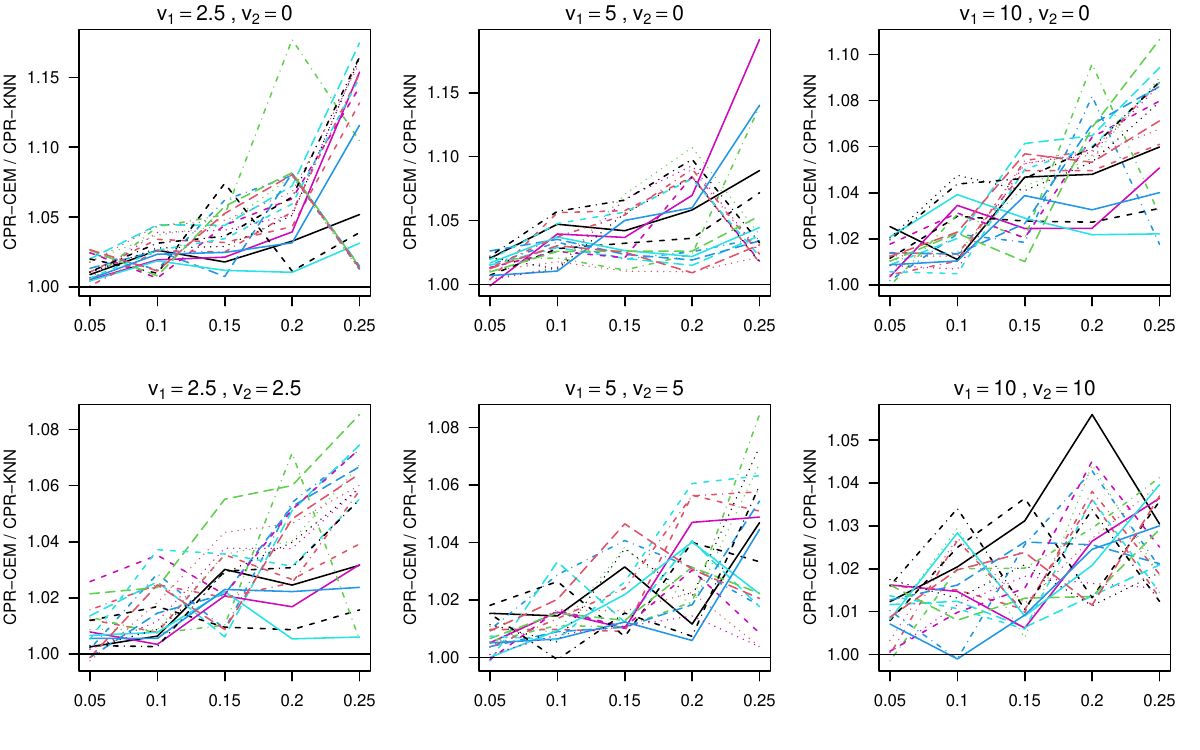} 
\includegraphics[width=.98\textwidth]{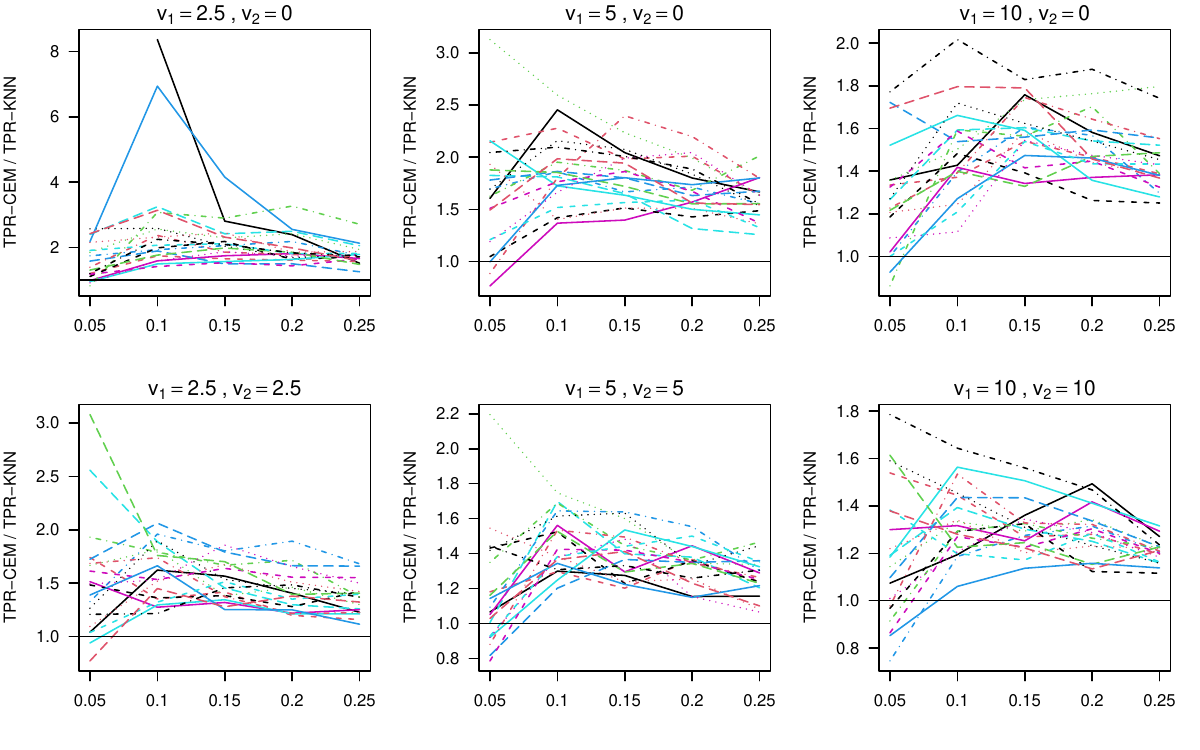} 
\includegraphics[width=.98\textwidth]{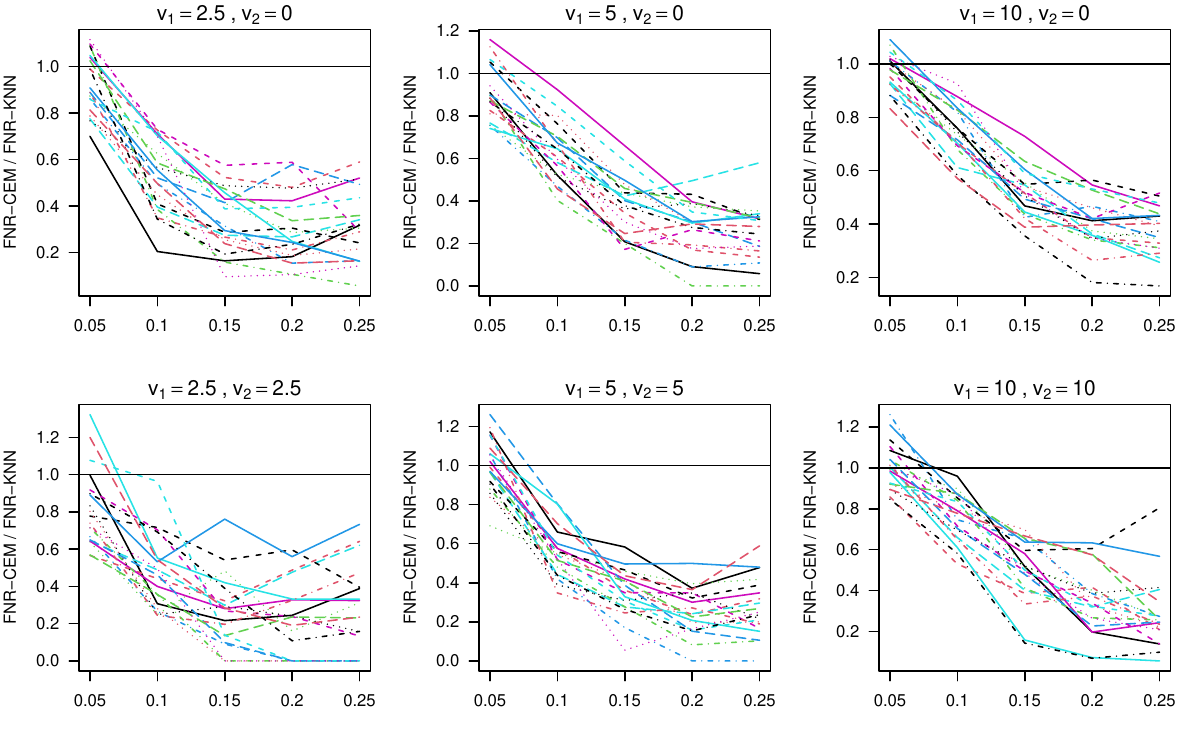} 
\end{center}	
\caption{\label{fig:ratio_adult} Iterated CEM vs KNN (\emph{adult} data): for $M=20$ simulated scenarios and threshold $q_D\in \{0.05, 0.1, 0.15, 0.2, 0.25\}$ the graphs show (i) the ratio of CEM and KNN correct prediction ratio (CPR), (ii) the ratio of CEM and KNN true positive ratio (TPR), and (iii) the ratio of CEM and KNN false negative ratio (FNR) (from top to bottom);  the percentage of $S=1$ (respectively, $S=0$) observations whose outcome is changed to $Y=0$ (respectively, $Y=1$) is denoted by $v_1$ (respectively, $v_2$).}
\end{figure}

\begin{figure}[h]
	\begin{center}
		\includegraphics[width=.98\textwidth]{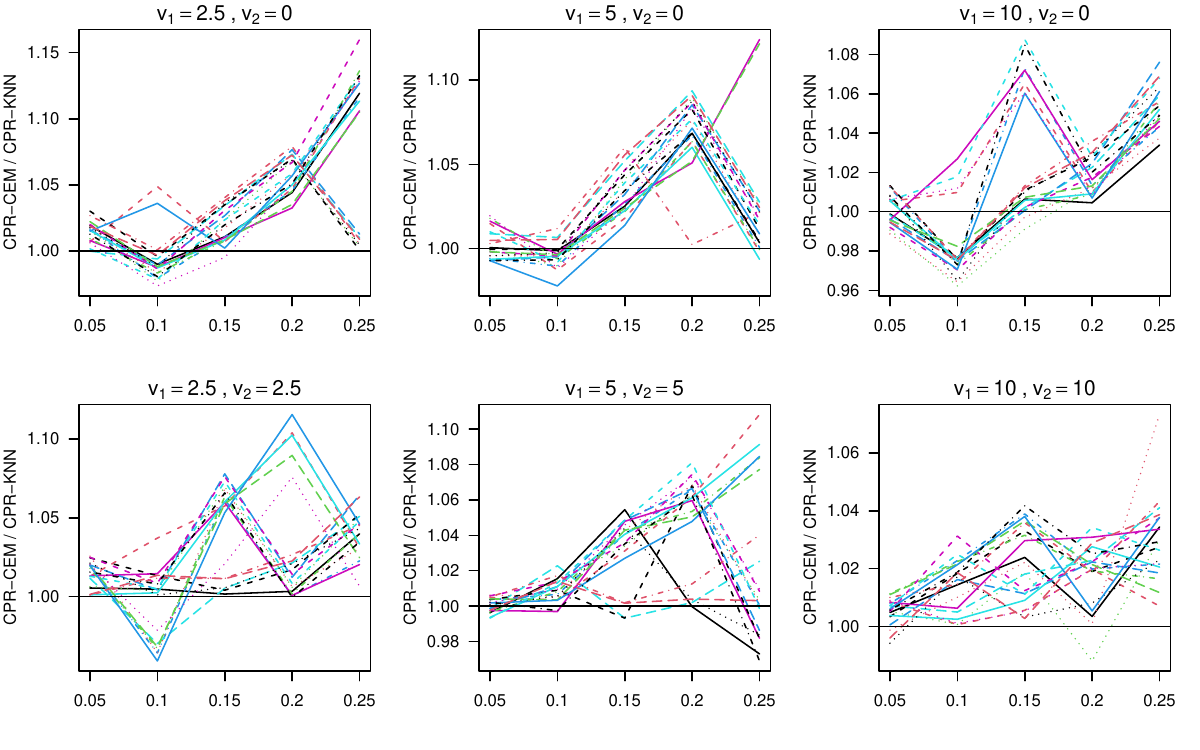} 
		\includegraphics[width=.98\textwidth]{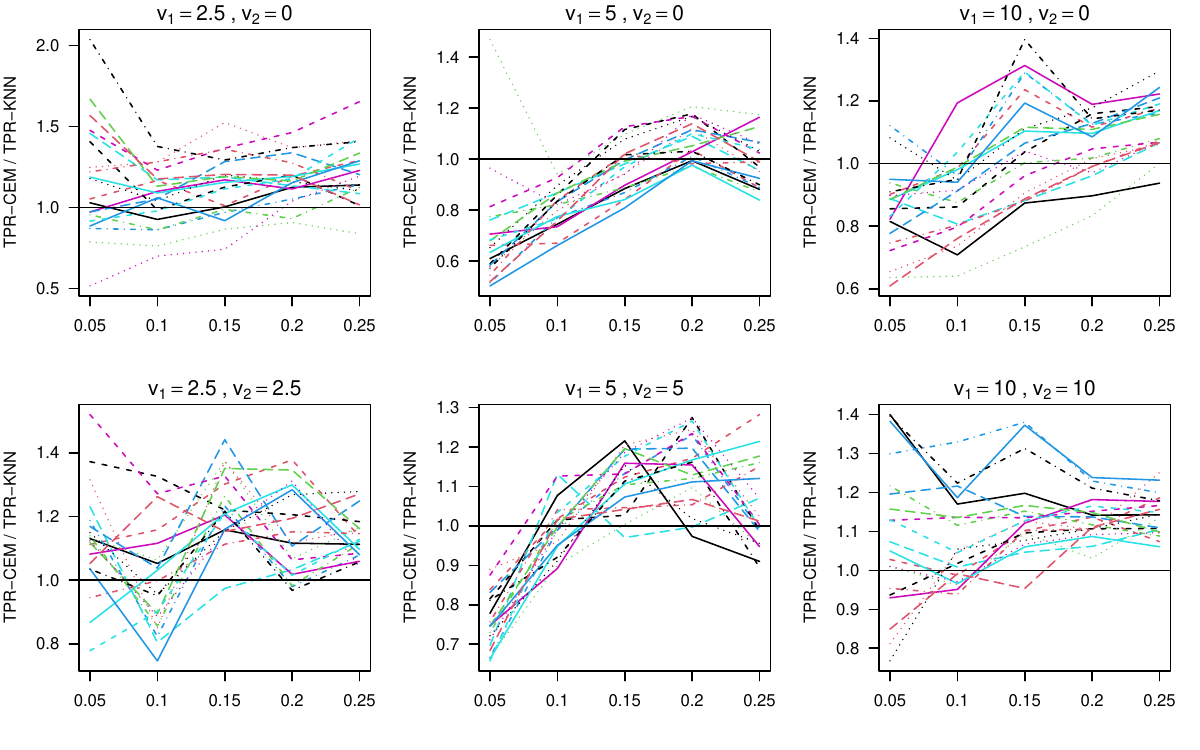} 
		\includegraphics[width=.98\textwidth]{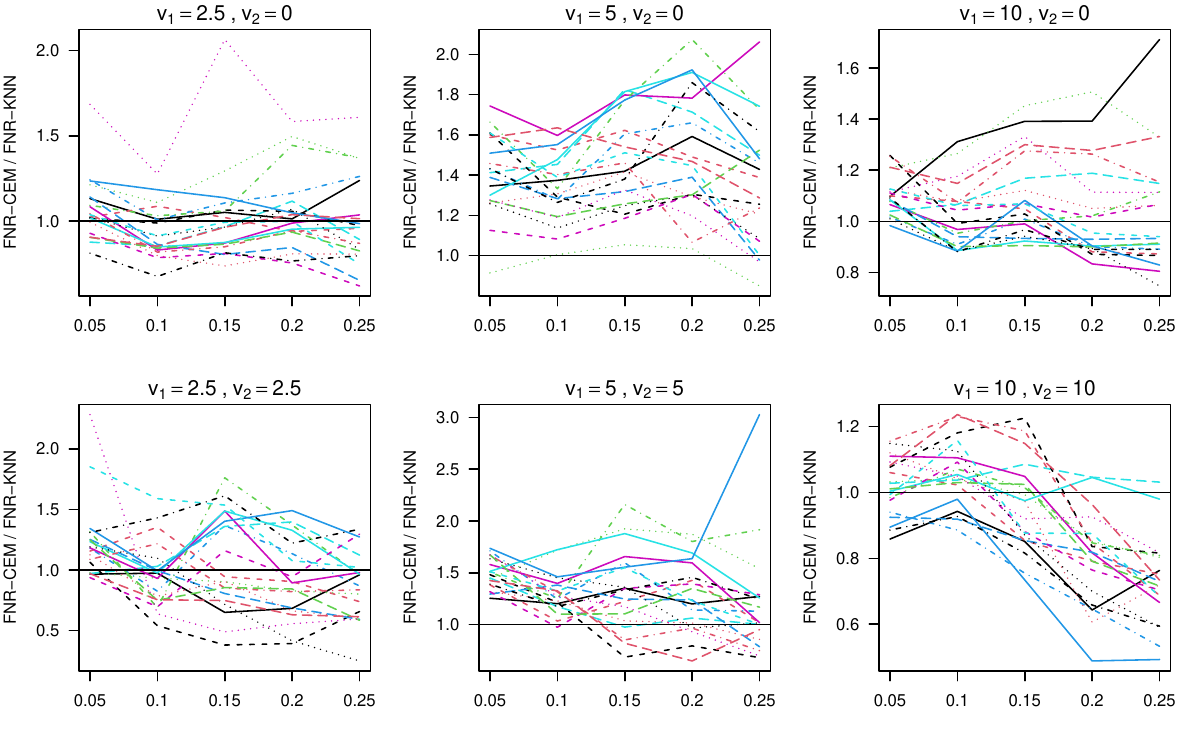} 
	\end{center}	
	\caption{\label{fig:ratio_compas} Iterated CEM vs KNN (\emph{COMPAS} data): for $M=20$ simulated scenarios and threshold $q_D\in \{0.05, 0.1, 0.15, 0.2, 0.25\}$ the graphs show (i) the ratio of CEM and KNN correct prediction ratio (CPR), (ii) the ratio of CEM and KNN true positive ratio (TPR), and (iii) the ratio of CEM and KNN false negative ratio (FNR) (from top to bottom);  the percentage of $S=1$ (respectively, $S=0$) observations whose outcome is changed to $Y=0$ (respectively, $Y=1$) is denoted by $v_1$ (respectively, $v_2$).}
\end{figure}

\begin{figure}[h]
	\begin{center}
		\includegraphics[width=.98\textwidth]{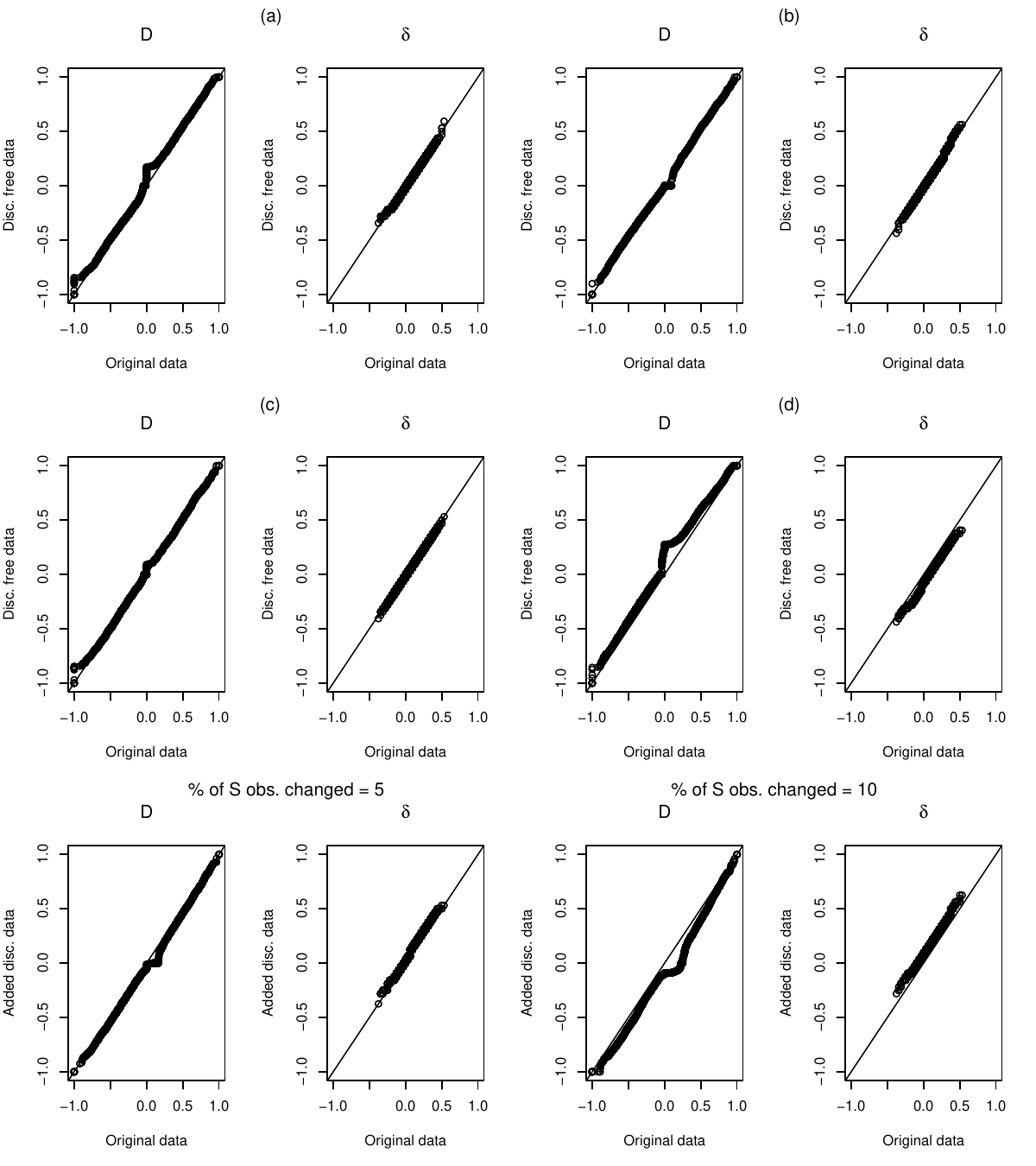} 
	\end{center}	
	\caption{\label{fig:qqplot_custody} Discrimination measures  performances for the \emph{Custody} data: qq-plots comparing the distributions of the discrimination scores $D$ and $\delta$ for (i) the discrimination free data against the original data according to strategies (a)--(d) described in Sect. \ref{sec:discadd} (first two rows from the top) and (ii) the data where discrimination has been introduced against the original data, by setting $p_1=p_2=5$  and  $p_1=p_2=10$ (bottom row).}
\end{figure}

\begin{figure}[h]
	\begin{center}
		\includegraphics[width=.98\textwidth]{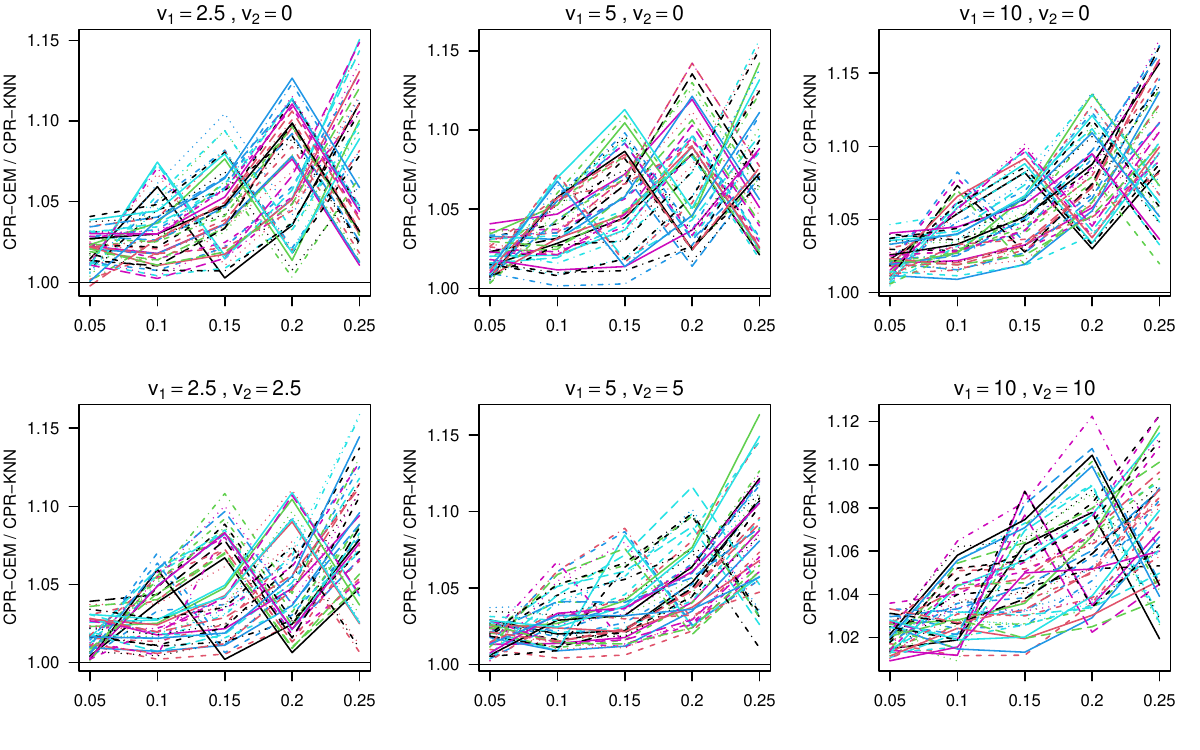} 
		\includegraphics[width=.98\textwidth]{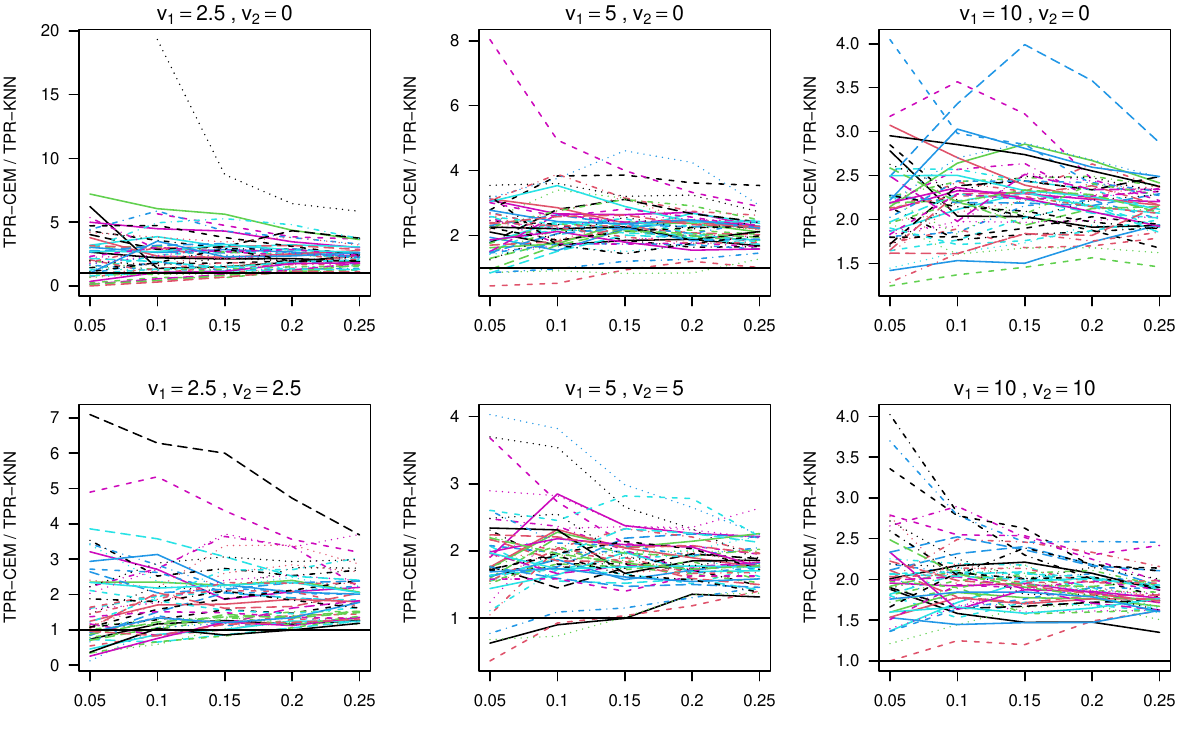} 
		\includegraphics[width=.98\textwidth]{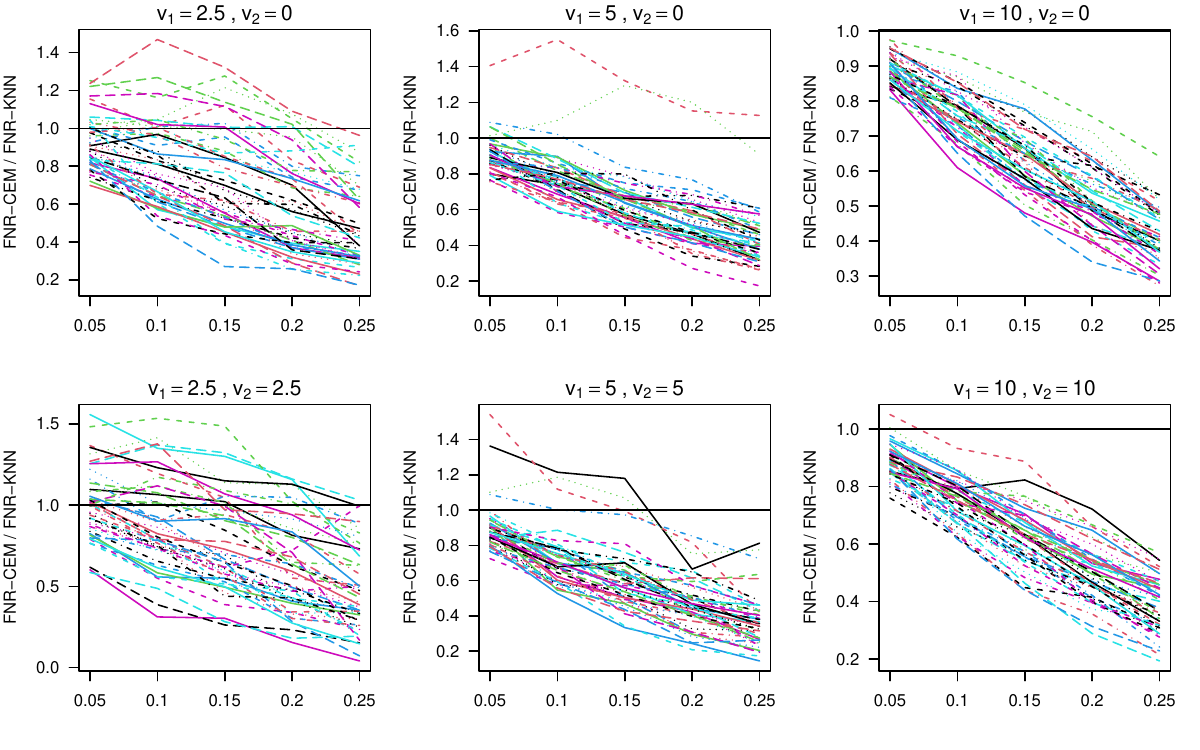} 
	\end{center}	
	\caption{\label{fig:ratio_custody} Iterated CEM vs KNN (\emph{Custody} data): for $M=20$ simulated scenarios and threshold $q_D\in \{0.05, 0.1, 0.15, 0.2, 0.25\}$ the graphs show (i) the ratio of CEM and KNN correct prediction ratio (CPR), (ii) the ratio of CEM and KNN true positive ratio (TPR), and (iii) the ratio of CEM and KNN false negative ratio (FNR) (from top to bottom);  the percentage of $S=1$ (respectively, $S=0$) observations whose outcome is changed to $Y=0$ (respectively, $Y=1$) is denoted by $v_1$ (respectively, $v_2$).}
\end{figure}

The CEM procedure and the $k$-nn approach are compared in terms of the performance of the classification (by a classification tree) by considering the correct prediction ratio, the true positive ratio, and the false negative ratio after changing the outcome of units with discrimination above a certain threshold $q_D$ for different combinations of $v_1$ and $v_2$ introduced above. In particular, we consider $(v_1, v_2)\in \{(2.5, 0), (5, 0), (10, 0), (2.5, 2.5), (5, 5),  (10, 10)\}$; for instance, if $(v_1, v_2)=(2.5, 0)$, then for the 2.5\% of the units in the data with $S=1$ the outcome becomes negative ($Y=0$), thus introducing discrimination against the protected category. Figures \ref{fig:ratio_adult}-\ref{fig:ratio_compas}-\ref{fig:ratio_custody} show the results for the \emph{adult}, \emph{COMPAS} and \emph{Custody} data, respectively; the threshold $q_D$ for discrimination detection corresponds to the percentile values $\{5, 10, 15, 20, 25\}$. The proposed measure shows an overall satisfactory performance compared to the competing method proposed by \cite{luong2011k}, yielding better results for the \emph{adult} dataset (with few exceptions at low quantiles, the values of correct prediction ratio and the true positive ratio are larger for CEM-based method than under the KNN approach, whereas the false negative ratio is smaller for CEM-based method than under the KNN approach).

\subsection{Introducing a conditioning variable correlated to S}

\begin{figure}[h]
	\begin{center}
    \includegraphics[width=\textwidth]{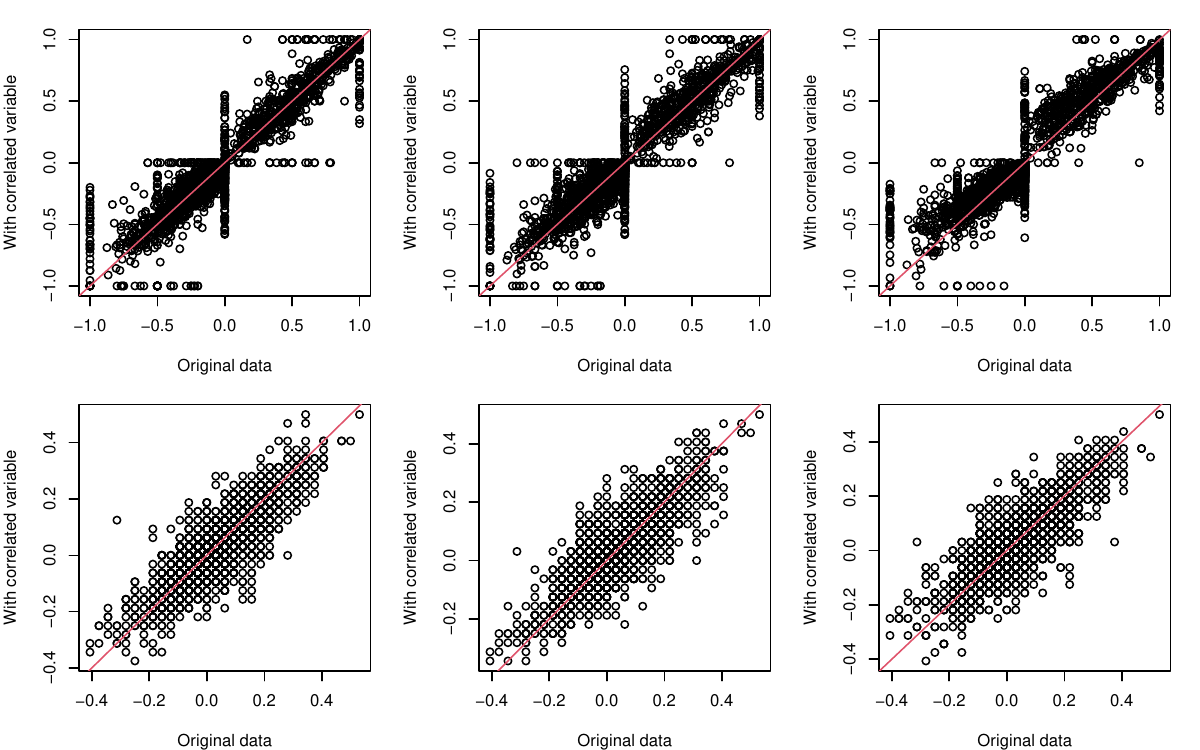} 
	\end{center}	
	\caption{\label{fig:corr_adult} Scatter plot of discrimination scores $D$ and $\delta$ for the \emph{adult} data (first and second row, respectively) estimated by the procedure by adding a variable correlated with $S$ (independent of the outcome), with correlation $\rho\in \{0.25, 0.5, 0.75\}$ (from left to right).}
\end{figure}

\begin{figure}[h]
	\begin{center}
		\includegraphics[width=\textwidth]{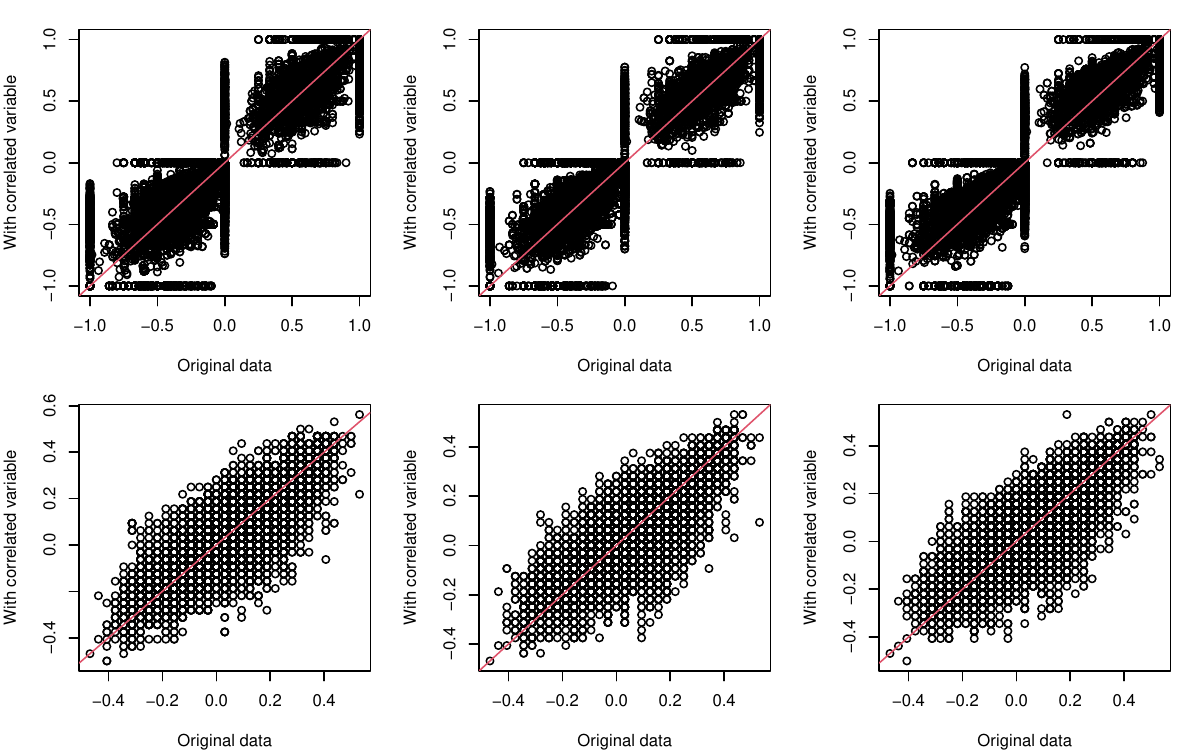} 
	\end{center}	
	\caption{\label{fig:corr_compas} Scatter plot of discrimination scores $D$ and $\delta$ for the \emph{COMPAS} data (first and second row, respectively) estimated by the procedure by adding a variable correlated with $S$ (independent of the outcome), with correlation $\rho\in \{0.25, 0.5, 0.75\}$ (from left to right).}
\end{figure}

\begin{figure}[h]
	\begin{center}
		\includegraphics[width=\textwidth]{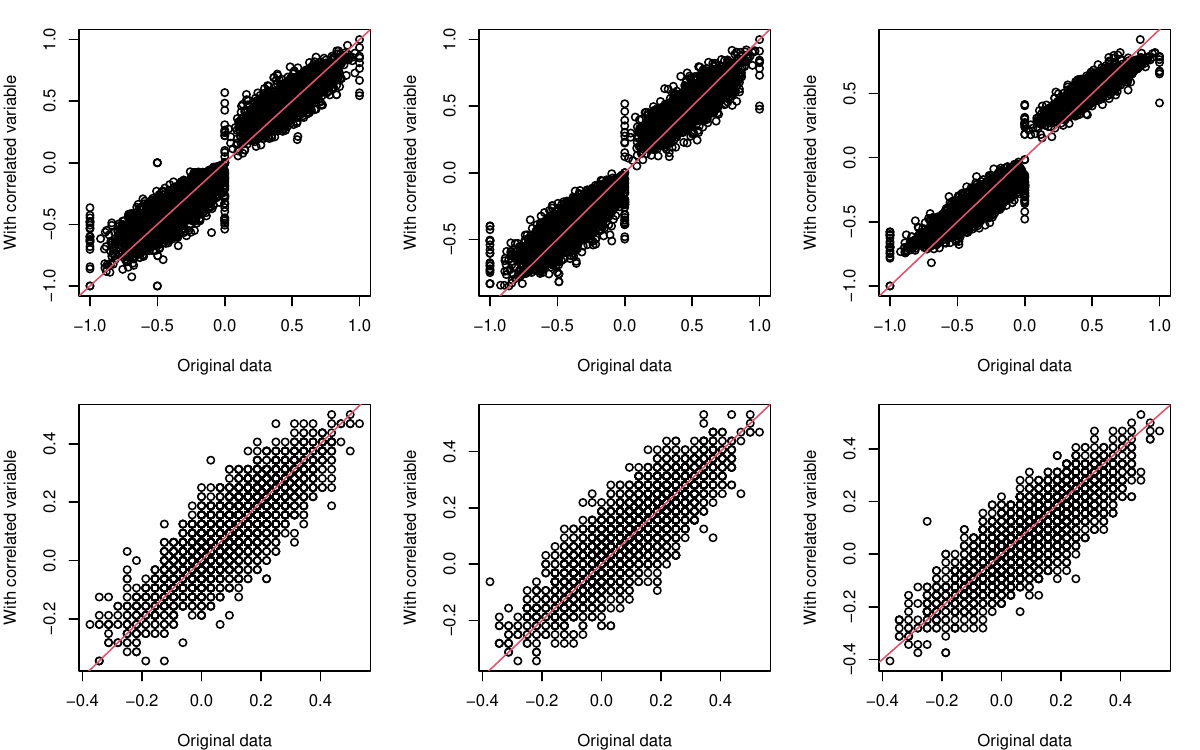} 
	\end{center}	
	\caption{\label{fig:corr_custody} Scatter plot of discrimination scores $D$ and $\delta$ for the \emph{Custody} data (first and second row, respectively) estimated by the procedure by adding a variable correlated with $S$ (independent of the outcome), with correlation $\rho\in \{0.25, 0.5, 0.75\}$ (from left to right).}
\end{figure}

In order to explore the extent to which the proposed measure is sensitive to discrimination against a protected group under different circumstances, we modify the original data by introducing a variable, $Z$, which is related to $S$ but unrelated to $Y$. In such situation, it is expected that the discrimination scores are lower when the variable related to the s.a.\ is included in the data, whereas they increase if such variable is omitted. In order to account for different degrees of correlation, we simulate $Z$ by adding a Gaussian noise with mean zero and variance computed in such a way that $Z$ and $S$ are correlated. Figures \ref{fig:corr_adult}-\ref{fig:corr_compas}-\ref{fig:corr_custody} display the scatter plots for the comparison between the $D_i$s and $\delta_i$s from the original data (\emph{adult}, \emph{COMPAS}, \emph{Custody}, respectively) and the data with $Z$ included, where the correlation is $\rho=0.25, 0.5, 0.75$. $D$ appears more sensitive to this. As expected, the discrimination is stronger if the variable is included, $\delta$ seems to be weakly affected by the presence of a variable correlated to $S$.
	
\section{Conclusions}
In order to detect inequality of treatment against protected classes in historical data it has been proposed to use the methods of causal inference to compare the treatment of statistical units belonging to the protected class to units not in the protected class which are similar with respect to the other observed characteristics. Similarity may be defined based on propensity scores or a distance metric. We argue that Coarsened Exact Matching (CEM) stratification could be more apt to this task since it allows matching of units only if their characteristics are equal (possibly after coarsening of numerical variables), rather than relying on an overall distance. Thus, we propose a measure of discrimination based on CEM stratification and consider different simulation scenarios to check the stability of the procedure and its performance in detecting and measuring discrimination against a protected group. Results suggest that the proposed measure is suitable to detect discrimination and in some cases outperforms the existing proposals.

\bibliographystyle{spmpsci}
\bibliography{pappadapaulireferencBIBTEX}
\end{document}